\documentclass[final,5p,times,twocolumn]{elsarticle}
\usepackage{framed}
\usepackage{multirow}
\usepackage{lineno}
\usepackage[dvipsnames,table]{xcolor}
\usepackage[utf8]{inputenc}
\usepackage{lineno}
\usepackage[hidelinks]{hyperref}
\usepackage{gensymb}
\usepackage{textcomp}
\usepackage{amsmath}
\usepackage{graphicx}
\usepackage{amsfonts}
\usepackage{booktabs}
\usepackage{subcaption}
\usepackage{soul}
\usepackage{multirow}
\usepackage{natbib}
\usepackage{color}
\usepackage{comment}
\usepackage{esvect}
\usepackage{nomencl} % Nomenclature package
    \makenomenclature
    \renewcommand*\nompreamble{\begin{multicols}{2}}
    \renewcommand*\nompostamble{\end{multicols}}
\usepackage[acronym,nomain,nonumberlist]{glossaries}
    \makeglossaries

\makeglossaries

\begin{document}

\begin{frontmatter}

\title{A Convex Hull Cheapest Insertion Heuristic for Precedence Constrained Traveling Salesperson Problems or Sequential Ordering Problems}
\tnotetext[label1]{This work was supported by Ford Motor Company through the Ford-OSU Alliance Program. Declarations of interest: none}

\author[mymainaddress]{Mithun Goutham\corref{mycorrespondingauthor}}
\cortext[mycorrespondingauthor]{Corresponding author}
\ead{goutham.1@osu.edu}

\author[mysecondaryaddress]{Meghna Menon}

\author[mysecondaryaddress]{Sarah Garrow}

\author[mymainaddress]{Stephanie Stockar}

\address[mymainaddress]{Department of Mechanical and Aerospace Engineering, Ohio State University, Columbus, OH 43210 USA}
\address[mysecondaryaddress]{The Ford Motor Company, Dearborn, MI 48126, USA}

% \thanks{Declarations of interest: none}

\begin{abstract}
The convex hull cheapest insertion heuristic is a well-known method that efficiently generates good solutions to the Traveling Salesperson Problem.
However, this heuristic has not been adapted to account for precedence constraints that restrict the order in which locations can be visited.
Such constraints result in the precedence constrained traveling salesperson problem or the sequential ordering problem, which are commonly encountered in applications where items have to be picked up before they are delivered.
In this paper, we present an adapted version of this heuristic that accounts for precedence constraints in the problem definition.
This algorithm is compared with the widely used Nearest Neighbor heuristic on the TSPLIB benchmark data with added precedence constraints.
It is seen that the proposed algorithm is particularly well suited to cases where delivery nodes are centrally positioned, with pickup nodes located in the periphery, outperforming the Nearest Neighbor algorithm in 97\% of the examined instances.
\end{abstract}

\begin{keyword}
Traveling salesman problems, Vehicle routing and navigation, Control and Scheduling, Algorithms, Logistics, Supply Chains
\end{keyword}

\end{frontmatter}

\section{Introduction}

Given a set of locations, the Traveling Salesperson Problem (TSP) finds the shortest possible tour that visits each location exactly once and returns to the starting location.
The TSP with Precedence Constraints (TSP-PC) \cite{mingozzi1997dynamic} or the Sequential Ordering Problem (SOP) \cite{escudero1988inexact} additionally has constraints that define precedence relations between locations, and a tour is feasible only when parent locations are visited before their respective children.
These problems arise in a variety of real-world applications where commodities have to be efficiently transported
% handling agents visit multiple locations in an order that minimizes the distance traveled 
while respecting precedence constraints related to picking up and dropping off items.
Applications seen in literature include freight transportation, crane scheduling in ports, job scheduling, automotive paint shops, helicopter scheduling and flexible manufacturing systems \cite{escudero1994lagrangian, ascheuer1996hamiltonian, ascheuer1993cutting, spieckermann2004sequential, fiala1992precedence}.

% \subsection{Approaches}
Heuristics that rapidly find reasonably good solutions are typically used when TSP solutions have to be found instantaneously \cite{xiang2008study, wong2014dynamic, markovic2015optimizing}, or to initialize exact methods for faster convergence to the optimal solution \cite{braekers2014exact, masmoudi2016three}.
This is because exact methods for computing the optimal solution to TSPs are intractable due to their $\mathcal{NP}$-hard nature \cite{jamal2017solving, shobaki2015exact,salii2019revisiting}. 
However, effective heuristics for the TSP-PC have been neglected in literature \cite{glover2001construction,taillard2022linearithmic}, and the simple Nearest Neighbor (NN) greedy heuristic is commonly used \cite{grigoryev2018solving, bai2020efficient, kumar2013traveling, nunes2016decentralized, edelkamp2018integrating}.

Among heuristics, the Convex Hull Cheapest Insertion (CHCI) heuristic has been shown to produce superior TSP solutions as compared to the NN heuristic which is frequently used because of its simplicity \cite{ivanova2021methods,warburton1993worst}.
The CHCI heuristic is initiated by a subtour created from the convex hull of points, and its interior points are then progressively added to the subtour in increasing order of insertion cost, until the complete tour is obtained.
The initiation of the candidate subtour with the convex hull of the TSP points is advantageous in generating good solutions because points on the boundary of the convex hull are visited in the same cyclic order as their order in the optimal TSP tour \cite{deineko1994convex, eilon1974distribution,golden1980approximate}.
However, the CHCI heuristic has not been adapted to the TSP-PC, where this property does not hold due to the added constraints.

The contribution of this paper is the extension of the CHCI algorithm to the TSP-PC, motivated by the expected reduction in tour cost when compared to the NN heuristic.
The proposed algorithm is referred to as the adapted CHCI (ACHCI) heuristic and its tours are initiated using the convex hull boundary points of only the pickup locations and the starting position.
For every ensuing insertion, precedence constraints are maintained by only permitting insertions into feasible segments of the subtour.
This implies that pickup locations can be added anywhere in the subtour, while delivery locations can only be added in the subtour segment that has already visited its corresponding pickup location. 
After describing the adapted approach with examples, the reduction in tour cost is compared to that of the NN heuristic, using TSPLIB benchmark instances that have been reproducibly modified for spatial and precedence constraints \cite{reinhelt2014tsplib}.
This paper shows experimentally that the algorithm outperforms the NN heuristic in 97\% of the cases when delivery locations are clustered around the centroid of the point cloud formed by all the locations of the TSP-PC.
% Taking into account all possible spatial configurations of the precedence constraints, it outperforms the Nearest Neighbor heuristic 69\% of the time. 

\section{Precedence Constrained TSP} \label{NE-TSP}

Consider a complete directed graph represented by $\mathcal{G}:=(V,A)$ where $V$ is the set of locations or nodes, and the directed or undirected arc set $A:=\{(v_i,v_j) | v_i,v_j \in V, \forall i\neq j \}$ connects every ordered pair of distinct locations in $V$.
A cost function $c : A \rightarrow \mathbb{R}^+$ defines the metric to be minimized, and each arc $(v_i,v_j) \in A$ is associated with a defined cost $c(v_i,v_j) \in \mathbb{R}^+$.
Let the arc costs be captured by a cost matrix $C\in\mathbb{R}^{n\times n}$ whose entries are defined by $C_{ij} = c(v_i,v_j) ~ \forall (v_i,v_j) \in A$.
% A Hamiltonian cycle, circuit or tour visits every node in the cost graph $G$ exactly once and then returns to the starting node.
% When $|V| = n$, the cycle may be defined as a sequence of arcs $P=(a_1,a_2,...,a_n)$, where the tail node of arc $a_1$ is also the head node of arc $a_n$, forming a graph cycle. 
The objective of the Traveling Salesperson Problem (TSP) is to find the minimum cost Hamiltonian tour on the graph $\mathcal{G}$, which is a sequence of consecutive arcs that visits every node in $V$ exactly once and returns to the starting node. 
Any Hamiltonian tour can be expressed as a sequence $T = (v_1,v_2, ... ,v_n,v_1)$ when $|V|=n$, and the tour cost is the sum of costs of constituting arcs given by $\textstyle J=\textstyle \sum_{r=1}^{n}c(v_r,v_{r+1})$ where $v_{n+1}=v_1$.

A precedence constraint, denoted by $v_i \prec v_j$, exists between nodes $v_i,v_j\in V$ if a path in the cost graph $\mathcal{G}$ is considered feasible only when node $v_i$ is visited at any point in the tour before $v_j$ is visited.
Such constraints induce a precedence directed graph $\mathcal{P}:=(V,R)$ where the existence of an arc $(v_i,v_j)\in R$ indicates $v_i \prec v_j$, and $v_i$ is called a parent node of the child $v_j$.
% The digraph $\mathcal{P}$ must be acyclic because precedence relations are transitively closed, i.e if $v_i\prec v_j,~v_j\prec v_k$, then $v_i\prec v_k$ \cite{balas1995precedence}.
To illustrate this, Fig. \ref{Im3a: PC} shows a set of parent and child nodes with precedence constraints illustrated using grey line segments, defining material handling tasks of moving items from the parent pickup positions, shown in red, to the respective children drop-off positions, marked in green. 
The moving agent starts and ends its tour at the depot, marked in blue.
The parent and child nodes along with the added depot location form the node set ${V}$ in both graphs $\mathcal{G}$ and $\mathcal{P}$, while the precedence relations feature as directional arcs that exist in $R$ of graph $\mathcal{P}$.
% The convex hull points of their 2D approximation is used to initiate the subtour as shown in Fig. \ref{Im3b: cHull subtour}.

With these precedence constraints, the objective of the TSP-PC or the SOP is to find a minimum cost Hamiltonian tour on the cost graph $\mathcal{G}$ while satisfying the arc constraints defined in graph $\mathcal{P}$.
Depending on the policies in place for the start and terminal states of the moving agent, a few variants in literature define the objective as a Hamiltonian path, instead of tour.
A traceable or Hamiltonian path visits every vertex in the cost graph $G$ exactly once, without returning to the start position.
Since the cost graph $\mathcal{G}$ is complete, 
any Hamiltonian path in $\mathcal{G}$ can be considered a Hamiltonian tour because an edge always exists that connects the start and end nodes.
Similarly, any Hamiltonian cycle in $\mathcal{G}$ can be converted to a Hamiltonian path by removing one of its edges.
Thus, each variant in literature can be posed as a minimum cost Hamiltonian tour problem 
and for this reason, without loss of generality, this paper presents a heuristic for the minimum cost Hamiltonian tour.
The discussion is also valid for Hamiltonian path versions of the problem.
% For this reason, the discussion in this paper is limited to the Hamiltonian cycle though the algorithm can be used for the following variants:
% The cycle starts and ends in the same vertex, but the path does not???.

\section{Adapted Convex Hull Cheapest Insertion Algorithm}

Let $v_0\in V$ be the start node or depot location with no predecessor or associated parent node in the precedence graph $\mathcal{P}$.
% , i.e. an empty set is defined by $\{v_i|(v_i,v_0)\in R\}$.
The set of all immediate successors of $v_0$ is denoted by the set $V_{0^+}:=\{v_j | (v_0,v_j)\in R\}$.
If $v_0$ has no child node, the set $V_{0^+}$ is comprised of only parent nodes.

\begin{figure}[t]
    \centering
  \subfloat[Definition of precedence constraints and depot \label{Im3a: PC}]{%
       \includegraphics[trim =0mm 0mm 0mm 0mm, clip, width=0.8\linewidth]{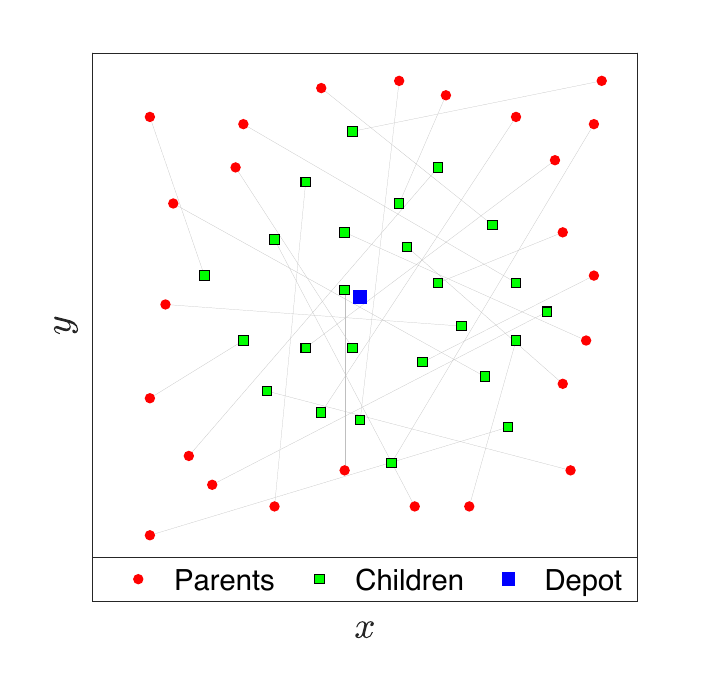}}
       \\
       \vspace{5mm}
  \subfloat[Initial convex hull subtour defined over parents and depot\label{Im3b: cHull subtour}]{%
        \includegraphics[trim =0mm 0mm 0mm 0mm, clip, width=0.8\linewidth]{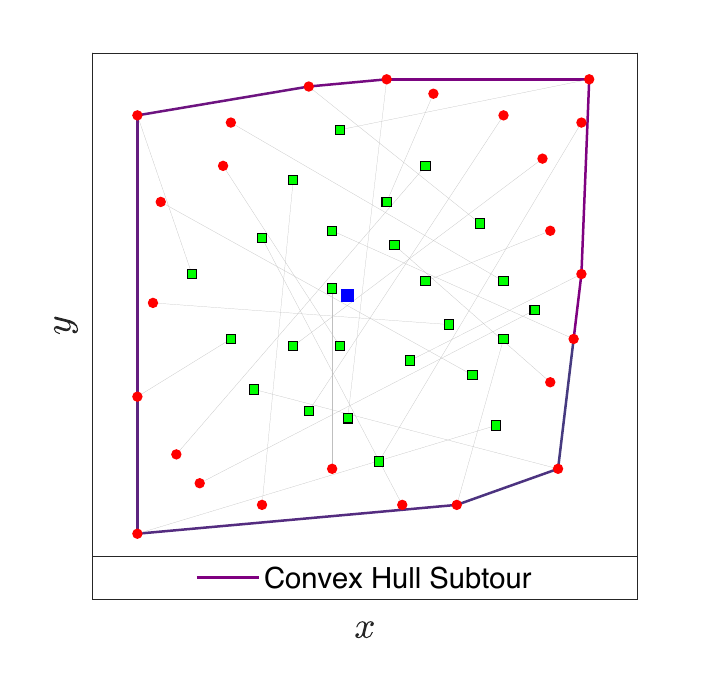}}
  \caption{ACHCI subtour initiation for the TSP-PC}
  \label{fig3} 
  \vspace{-2mm}
\end{figure}

The algorithm is initiated by obtaining the ordered boundary nodes of the convex hull of set $\bar{V}_0:=V_{0^+} \cup \{v_0\}$.
This ordered set of points is denoted by $T_0$, but because a notion of precedence exists, $T_0$ is assigned a direction for traversal before inserting the remaining nodes in $V \setminus T_0$.
The convex hull $T_0$ drawn over the points $\bar{V}_0$, as shown in Fig.\ref{Im3b: cHull subtour}, should therefore be first assigned a direction to be considered a subtour.
% $\overset{\leftarrow}{v}\overset{\rightarrow}{T}$
For a chosen clockwise or anti-clockwise direction, denote the subtour as $\overleftarrow{T}$.
The subsequent steps are repeated for the other direction $\overrightarrow{T}$.

Let the node sequence $[v_1, v_2, ..., v_1]$ form the initializing subtour $\overleftarrow{T}$.
For some node $v_p\in \overleftarrow{T}$, let the segment of $\overleftarrow{T}$ that has already visited $v_p$ be denoted by $\overleftarrow{T}_{p^+} := [v_p,v_{p+1},..., v_1]$.
For a candidate insertion node $v_k \notin \overleftarrow{T}$, the feasible partition of $\overleftarrow{T}$ where $v_k$ can be inserted, is the subtour segment that contains every parent of this candidate.
If $V_{k}^-:=\{v_i | (v_i,v_k)\in R\}$ is the set of all parents of $v_k$, then, the insertion of a candidate node $v_k \notin \overleftarrow{T}$ is only permitted in $\overleftarrow{T}^k = \bigcap_{p\in V_{k}^-}\overleftarrow{T}_{p^+}$.

\begin{figure}[t]
    \centering
  \subfloat[The first instance when a child node is inserted onto the valid segment (highlighted) of the counterclockwise tour \label{Im4a: Insert1 CCW}]{%
       \includegraphics[trim =0mm 0mm 0mm 0mm, clip, width=0.8\linewidth]{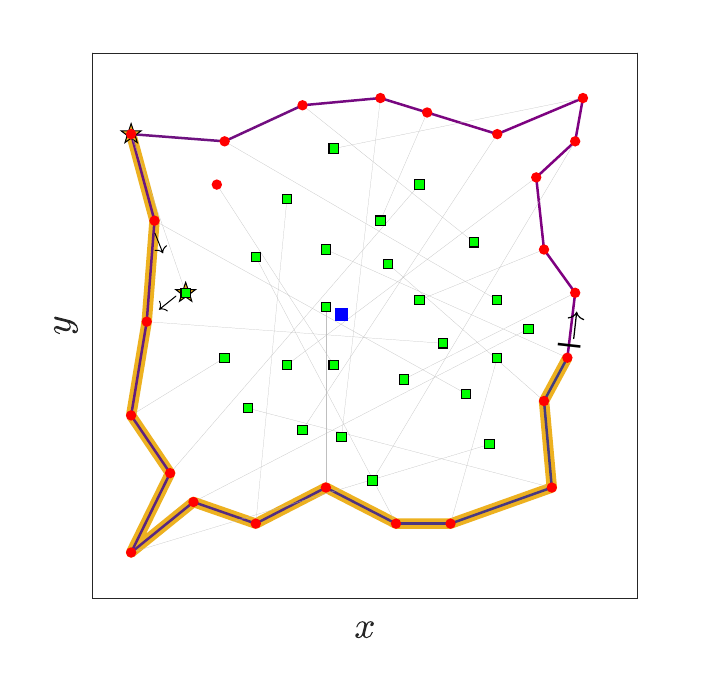}}
       \\
       \vspace{6mm}
  \subfloat[Completed counterclockwise ACHCI tour that maintains precedence constraints \label{Im4b: Tour CCW}]{%
        \includegraphics[trim =0mm 0mm 0mm 0mm, clip, width=0.8\linewidth]{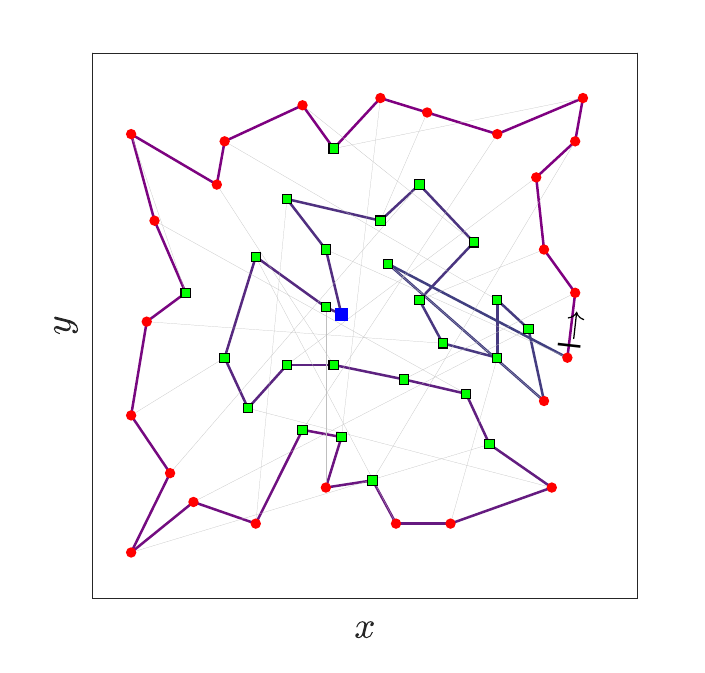}}
  \caption{Illustrative example tour of the ACHCI algorithm}
  \label{fig2} 
  \vspace{-2mm}
\end{figure}

For every node $v_i \notin \overleftarrow{T}$ that is yet to be inserted to the subtour, the insertion arc given by consecutive nodes $(v_q,v_r) \in \overleftarrow{T}^i$ is found that minimizes insertion cost ratio $(C_{qi} + C_{ir})/C_{qr}$.
To ensure feasibility, child nodes whose parents have not yet been visited in $\overleftarrow{T}$ are assigned an infinite insertion cost because they have an empty feasible partition.
% Every node $v_j \notin \overleftarrow{T}$ with an empty feasible partition $\overleftarrow{T}^j$ is assigned an infinite insertion cost.
Next, the node $v^* \notin \overleftarrow{T}$ with the lowest insertion cost ratio is inserted at its associated insertion arc.
This increments $\overleftarrow{T}$ by one node, and these steps are repeated until all the nodes have been inserted.

\begin{figure*}[t!]
\centering
  \subfloat[TSPLIB instance `eil76' with central children \label{a: cChildren}]{%
       \includegraphics[trim =0mm 0mm 0mm 0mm, clip, width=0.3\linewidth]{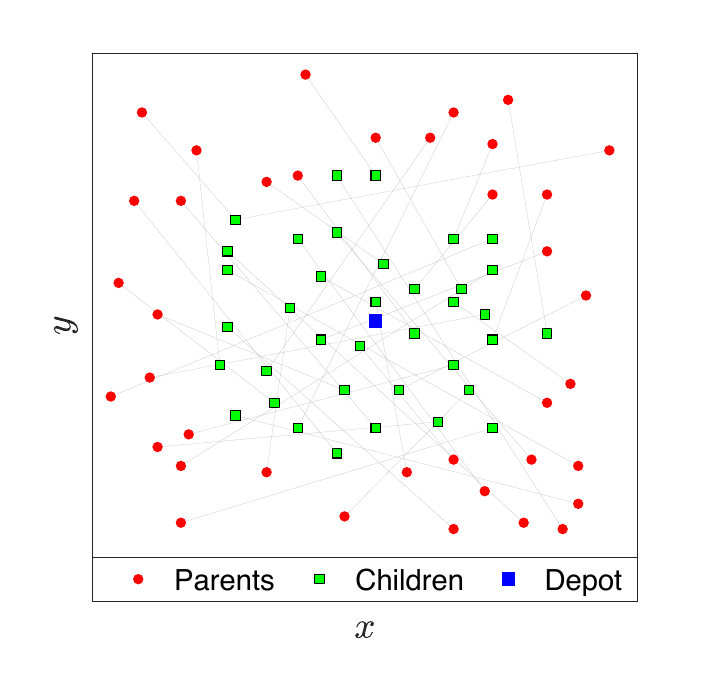}}
\hfill
  \subfloat[ TSPLIB instance `eil76' with central parents\label{b: cParents}]{%
        \includegraphics[trim =0mm 0mm 0mm 0mm, clip, width=0.3\linewidth]{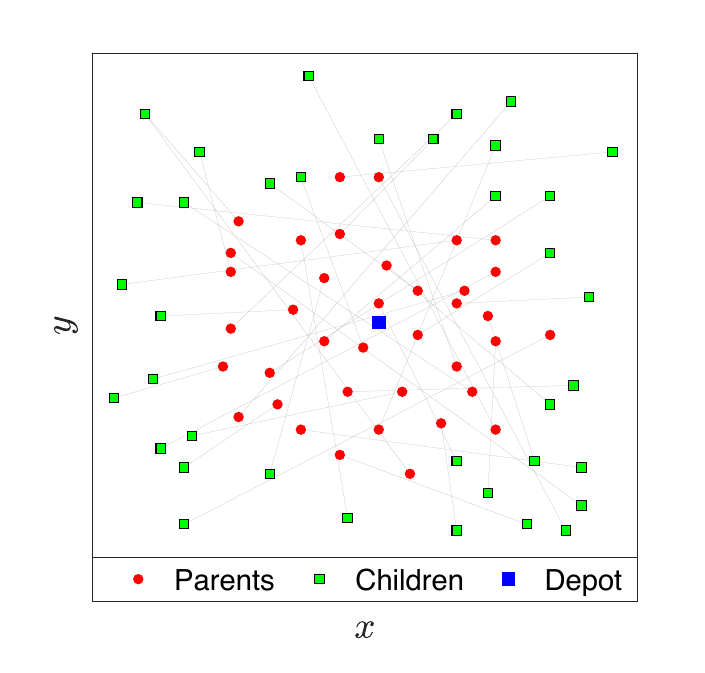}}
\hfill
  \subfloat[ TSPLIB instance `eil76' with random precedence\label{c: rPrecedence}]{%
        \includegraphics[trim =0mm 0mm 0mm 0mm, clip, width=0.3\linewidth]{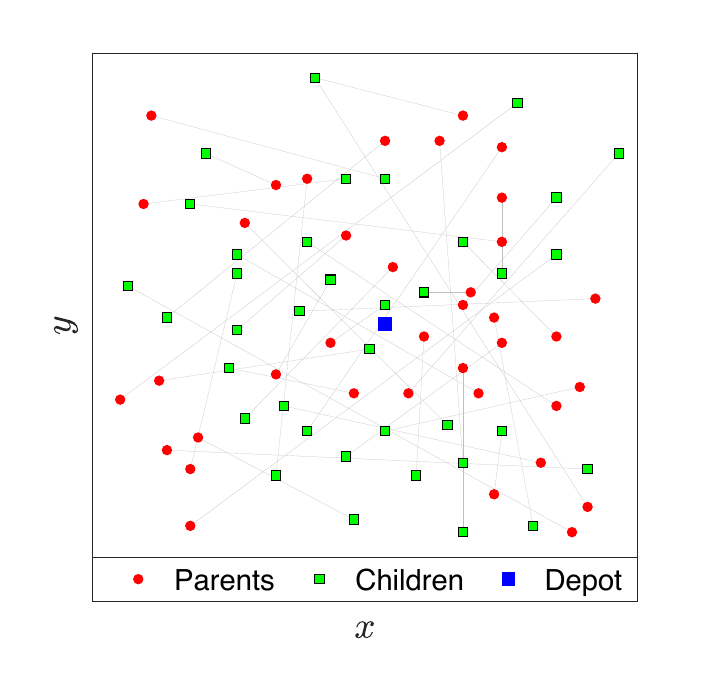}}\vspace{5mm}\\
\hspace{1mm}      
  \subfloat[Cost ratio histogram for central children \label{d: resChildren}]{%
       \includegraphics[trim =0mm 0mm 0mm 0mm, clip, width=0.27\linewidth]{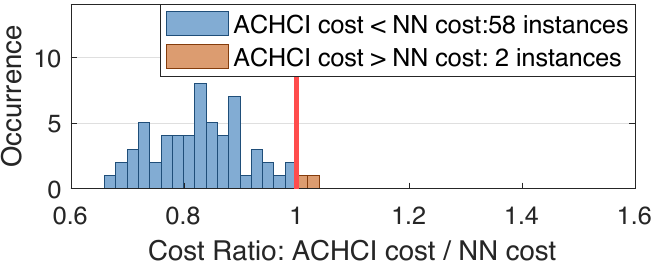}}
\hspace{13.5mm} 
  \subfloat[Cost ratio histogram for central parents\label{e: resParents  }]{%
        \includegraphics[trim =0mm 0mm 0mm 0mm, clip, width=0.27\linewidth]{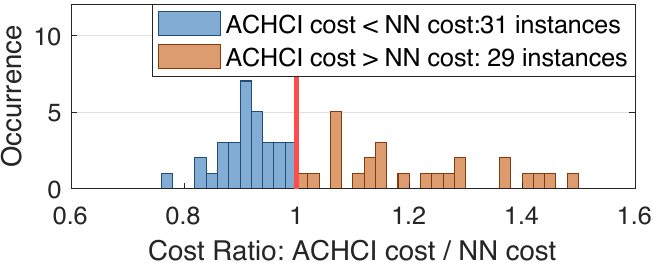}}
\hspace{13.5mm} 
  \subfloat[Cost ratio histogram for random precedence\label{f: resRandom}]{%
        \includegraphics[trim =0mm 0mm 0mm 0mm, clip, width=0.27\linewidth]{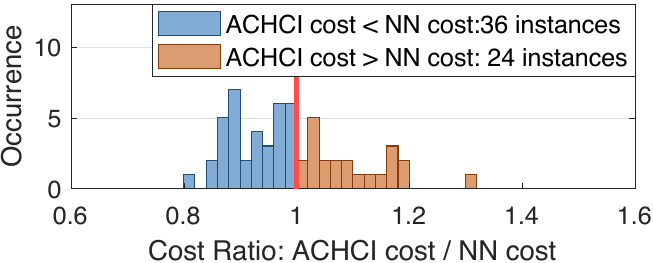}}
        \hspace{2mm}
  \caption{Performance comparison of the ACHCI and NN algorithm for various spatial configurations of precedence constraints}
  \label{PC-examples2} 
\end{figure*}

\begin{table*}[t]
\centering
\caption{Percentage reduction of NN heuristic cost when using the ACHCI heuristic}\label{table1}
\begin{minipage}[t]{0.4\textwidth}
\centering
    \begin{tabular}[t]{rccc}
    \toprule
    \begin{tabular}[c]{@{}l@{}}TSPLIB \\ instance\end{tabular} & \begin{tabular}[c]{@{}l@{}}Central \\ Children\end{tabular} & \begin{tabular}[c]{@{}l@{}}Central \\ Parents\end{tabular} & \begin{tabular}[c]{@{}l@{}}Random\\ Precedence\end{tabular} \\
    \midrule
\textit{eil51}    &15.5   &10.4   &11.5   \\ 
\textit{t70}    & 5.7   & 2.7   &10.5   \\ 
\textit{eil76}    &17.1   & 0.5   &10.1   \\ 
\textit{erlin52}    &17.8   & 8.9   & 0.8   \\ 
\textit{eil101}    &26.8   &14.3   &11.1   \\ 
\textit{rat99}    &23.3   &\textbf{-6.7}   &\textbf{-3.5}   \\ 
\textit{pr76}    &22.2   &\textbf{-12.0}   &14.8   \\ 
\textit{roC100}    &17.0   &\textbf{-29.4}   &19.4   \\ 
\textit{roD100}    &11.9   &\textbf{-37.0}   & 1.8   \\ 
\textit{roE100}    &15.8   &\textbf{-23.4}   & 7.2   \\ 
\textit{roA100}    & 1.0   & 7.5   &11.1   \\ 
\textit{roB100}    &20.3   &\textbf{-6.1}   & 4.0   \\ 
\textit{in105}    &29.8   & 4.7   &\textbf{-16.7}   \\ 
\textit{pr107}    &11.3   &\textbf{-14.3}   &\textbf{-19.6}   \\ 
\textit{pr124}    &24.8   &\textbf{-0.2}   & 8.5   \\ 
\textit{roB150}    &19.8   &\textbf{-14.9}   & 0.1   \\ 
\textit{roA150}    &14.2   &\textbf{-0.5}   &\textbf{-11.3}   \\ 
\textit{pr136}    &27.2   &13.0   &11.3   \\ 
\textit{pr144}    & 0.4   &\textbf{-41.1}   &\textbf{-0.3}   \\ 
\textit{pr152}    & 7.5   &\textbf{-44.6}   &\textbf{-7.8}   \\ 
\textit{rat195}    &11.0   &\textbf{-13.6}   & 3.0   \\ 
\textit{bier127}    &27.5   &17.5   & 3.7   \\ 
\textit{roA200}    &24.2   & 6.9   & 1.7   \\ 
\textit{roB200}    & 9.0   &\textbf{-29.1}   &\textbf{-19.4}   \\ 
\textit{rd100}    &32.0   & 8.8   &12.0   \\ 
\textit{gil262}    &23.3   &\textbf{-7.0}   & 1.3   \\ 
\textit{pr226}    & 2.8   &\textbf{-43.9}   &\textbf{-16.3}   \\ 
\textit{a280}    &19.2   &\textbf{-1.3}   & 7.7   \\ 
\textit{ts225}    &19.0   &\textbf{-26.4}   &\textbf{-13.7}   \\ 
\textit{pr264}    &\textbf{-1.8}   &\textbf{-24.2}   &\textbf{-9.0}   \\ 
    \bottomrule
    \end{tabular}
\end{minipage}% 
\hspace{8mm}
\begin{minipage}[t]{0.4\textwidth}
\centering
    \begin{tabular}[t]{rccc}
    \toprule
    \begin{tabular}[c]{@{}l@{}}TSPLIB \\ instance\end{tabular} & \begin{tabular}[c]{@{}l@{}}Central \\ Children\end{tabular} & \begin{tabular}[c]{@{}l@{}}Central \\ Parents\end{tabular} & \begin{tabular}[c]{@{}l@{}}Random\\ Precedence\end{tabular} \\
    \midrule
\textit{tsp225}    &12.6   &\textbf{-14.9}   &14.7   \\ 
\textit{pr299}    &16.1   & 9.9   &\textbf{-15.7}   \\ 
\textit{in318}    &11.2   & 7.6   & 4.3   \\ 
\textit{in318}    &11.2   & 7.6   &\textbf{-3.9}   \\ 
\textit{h130}    & 7.3   & 8.1   &\textbf{-5.0}   \\ 
\textit{u159}    &14.0   &\textbf{-0.2}   &\textbf{-2.5}   \\ 
\textit{h150}    & 7.9   & 1.0   & 6.6   \\ 
\textit{d198}    &29.8   &13.3   &13.7   \\ 
\textit{pr439}    &21.4   & 1.1   & 5.2   \\ 
\textit{rat575}    &15.2   &\textbf{-6.1}   & 8.3   \\ 
\textit{rat783}    &27.1   &\textbf{-6.9}   &\textbf{-6.0}   \\ 
\textit{rd400}    &14.7   &10.6   &\textbf{-4.1}   \\ 
\textit{fl417}    &\textbf{-3.6}   &\textbf{-49.7}   &\textbf{-30.2}   \\ 
\textit{pcb442}    &12.7   &23.4   &\textbf{-9.8}   \\ 
\textit{d493}    &19.9   &17.0   & 7.6   \\ 
\textit{pr1002}    &21.3   & 3.1   &10.6   \\ 
\textit{u574}    &16.9   & 3.5   &\textbf{-1.9}   \\ 
\textit{p654}    &11.5   &\textbf{-36.4}   &\textbf{-3.5}   \\ 
\textit{d657}    &10.9   & 9.9   &13.4   \\ 
\textit{u724}    &17.1   &\textbf{-19.7}   &13.1   \\ 
\textit{u1060}    &16.4   &\textbf{-12.4}   & 3.8   \\ 
\textit{vm1084}    &20.9   &\textbf{-0.7}   & 3.6   \\ 
\textit{nrw1379}    &13.7   &10.5   & 5.6   \\ 
\textit{pcb1173}    &29.7   & 9.1   & 0.7   \\ 
\textit{d1291}    & 6.0   & 4.7   &13.7   \\ 
\textit{rl1304}    &23.0   & 7.7   &\textbf{-1.2}   \\ 
\textit{rl1323}    &31.4   & 4.3   & 2.6   \\ 
\textit{fl1400}    &30.7   &12.0   &\textbf{-3.6}   \\ 
\textit{u1432}    &17.2   & 9.6   &\textbf{-16.5}   \\ 
\textit{fl1577}    &27.7   &\textbf{-3.7}   &\textbf{-0.7}   \\  
    \bottomrule
    \end{tabular}
\end{minipage}
\end{table*}

Considering the initial subtour $T_0$ was assigned some arbitrary direction that resulted in $\overleftarrow{T}$, all of the tour building steps are also repeated after initializing subtour $T_0$ in the opposite direction, forming another subtour $\overrightarrow{T}$.
Thus, two complete tours $\overleftarrow{T}$ and $\overrightarrow{T}$ are obtained for the TSP-PC, and the minimum cost tour is selected as the heuristic.

In summary, the adapted CHCI (ACHCI) algorithm for the TSP-PC is as follows:
\begin{enumerate}[\textit{Step} 1:]
    \item Initiate subtour $T_0$ as the ordered nodes of the convex hull of the start node and parent nodes.
    \item Assign a direction to subtour $T_0$, name it subtour $\overleftarrow{T}$.
    \item Find $v_k$ not in subtour $\overleftarrow{T}$, and consecutive nodes $v_i,v_j$ in its insertable segment that minimizes insertion cost ratio ($C_{ik} + C_{kj})/C_{ij}$.
    % \item For every $v_k$ not in subtour 1, find consecutive nodes $v_i,v_j$ in the valid partition of the subtour that minimizes insertion cost $C_{ik} + C_{kj} - C_{ij}$, obtained from the cost matrix $C$. 
    \item Increment subtour $\overleftarrow{T}$ by inserting $v_k$ between $v_i$ and $v_j$.
    \item Repeat \textit{Step} 3 and \textit{Step} 4, to obtain the complete Hamiltonian cycle $\overleftarrow{T}$ of tour cost $\overleftarrow{J}$.
    \item Assign the other direction to subtour $T_0$, name it subtour $\overrightarrow{T}$.
    \item Repeat \textit{Step} 3 and \textit{Step} 4, but on $\overrightarrow{T}$, to obtain another Hamiltonian cycle $\overrightarrow{T}$ of tour cost $\overrightarrow{J}$.
    \item Choose the minimum cost tour between $\overrightarrow{T}$ \& $\overleftarrow{T}$.
\end{enumerate}

For the convex hull points shown in Fig. \ref{Im3b: cHull subtour}, the tour with the counterclockwise direction is shown in Fig. \ref{Im4a: Insert1 CCW} at the first instance when a child node, marked by a green star, is inserted.
Its associated parent is marked by the red star, and the valid partition of the tour is highlighted, being the segment that has already visited the parent.
Notice also that at this instance, the sub-tour $\overrightarrow{T}$ has already been incremented by other parent nodes, but not all parent nodes have been inserted.
After inserting all the remaining nodes, the completed Hamiltonian tour with some cost $\overrightarrow{J}$ is shown in Fig. \ref{Im4b: Tour CCW}.
It can be verified that every node is visited while respecting the defined precedence constraints.
These steps are also repeated after initiating $T_0$ in the clockwise direction, to obtain a tour of some cost $\overleftarrow{J}$. 
The tour with lower cost between $\overrightarrow{J}$ \& $\overleftarrow{J}$ is selected as the ACHCI heuristic tour.

\section{Nearest Neighbor Heuristic}
The benchmark for the proposed ACHCI algorithm is the Nearest Neighbor (NN) which is known experimentally to perform reasonably well \cite{charikar1997constrained} and is commonly seen in constrained TSP literature \cite{grigoryev2018solving,bai2020efficient,kumar2013traveling,nunes2016decentralized,edelkamp2018integrating}.
It is a fast and simple greedy selection rule \cite{dacey1960selection} that can easily be modified to account for constraints in the problem formulation.
Starting from a defined or randomly selected initial node, the NN algorithm assigns the nearest unvisited node as the next node until all nodes are contained in the path.
By modifying this slightly to only consider unvisited feasible nodes as the next node, it can be adapted for precedence-constraints, as follows:
\begin{enumerate}[\textit{Step} 1:]
    \item Initiate the subtour as the depot
    \item Add the closest feasible node to the end of the subtour
    \item Repeat step 2 until all the nodes are included
    \item Return to the depot
\end{enumerate}

\section{Computational Experiments}

To compare the effectiveness of the ACHCI algorithms with the simple NN algorithm, sufficiently diverse benchmark instances are not readily available for the TSP-PC.
For this reason, the popular TSPLIB benchmark instances \cite{reinhelt2014tsplib} are modified in a reproducible manner, to add precedence and environmental constraints. 

\begin{enumerate}[\textit{Step} 1:]
    \item Load a TSPLIB point cloud that has 2D Cartesian coordinates defined for every point. Let $n$ be the number of points.
    \item Find the centroid of the point cloud.
    \item Sort and assign indices to the points in order of increasing distance from the centroid. Let the indices be $1,2,...,n$ for the sorted points.
    \item Designate the point with index 1 as the depot or starting node. It has no parents.
    \item Define precedence constraints between points with pairs of indices as $(n\prec 2),(n-1\prec 3),(n-2\prec 4)$ and so on. If only three nodes remain to be assigned precedence constraints, the node with the smallest index is assigned to be the child node while the other two are its parents.
\end{enumerate}
\begin{figure}[b]
    \centering
        \includegraphics[trim =0mm 0mm 0mm 0mm, clip, width=0.8\linewidth]{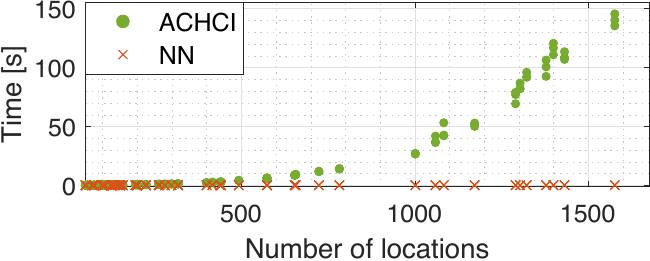}
  \caption{Computation time taken by the ACHCI and NN heuristics}
  \label{4: PC-time} 
\end{figure}

The generated point cloud has child nodes that are clustered closer to the centroid, with parent nodes at the periphery.
By changing the direction of the precedence constraints in \textit{Step 5}, the complementary case is produced, where parent nodes are closer to the centroid.
Another case with randomly assigned precedence constraints is also studied.
These three cases of precedence constraints are illustrated using the modified TSPLIB `eil51' instance in Fig. \ref{a: cChildren}, \ref{b: cParents} and \ref{c: rPrecedence}.

The experiments are conducted for 60 TSPLIB instances, with the number of points varying from 51 to 1,577.
The results of these experiments are summarized in Table \ref{table1}, where the first column lists the name of each TSPLIB instance, formatted with a prefix followed by a numeric value that indicates the number of points in the respective instance.
The remaining columns provide the percentage reduction in NN tour cost when using the ACHCI heuristic for the three types of spatial variations in precedence constraints.

Since child nodes can only be inserted in feasible segments of the tour, their spatial locations significantly affect the performance of the ACHCI algorithm.
The histogram in Fig. \ref{d: resChildren} shows that the ACHCI tour cost is lower than the NN cost in 97\% of the cases where children nodes are spatially closer to the centroid.
This is because when parent nodes are located toward the periphery, future insertions onto the resulting convex hull are restricted to be within the region enclosed by the tour and large insertion segments are avoided.

When parent nodes are centrally located however, the insertions of child nodes from the periphery can result in large deviations from the tour depending on the distance of the child node from its feasible insertion segment of the subtour.
This behavior is most significant when only a few child nodes are left to be inserted, but their insertable subtour segments happen to be far from them.
As a result, the ACHCI algorithm does not perform as well in such cases, as seen in Fig. \ref{e: resParents  }.
For randomly located precedence constraints, the ACHCI tour cost is lower than the NN tour in 60\% of the test cases as seen in Fig. \ref{f: resRandom}.

These experiments were conducted in a Matlab R2023b environment on an AMD Ryzen 5600X CPU clocked at 3.7 GHz.
The computation times taken by the two heuristics are shown in Fig. \ref{4: PC-time}, showing a worst case complexity of $O(n^3)$ which is characteristic of the cheapest first insertion criteria.
When compared to the ACHCI algorithm, the NN heuristic almost instantaneously provides good tours regardless of the number of points.
Thus, depending on the use-case and computation time budget available, it may be worthwhile to compute the ACHCI tour in addition to the NN tour and simply choose the tour with minimum cost between the two.
Further, because the ACHCI algorithm for the TSP-PC considers the initiated subtour in two directions, it is well suited for parallel computing to reduce computation time.

\section{Conclusion}

The well-known CHCI algorithm has been adapted for the precedence constrained generalization of the TSP.
The convex hull is drawn over only the non-child nodes when initiating the ACHCI algorithm, and the subsequent insertions are permitted in segments that respect the precedence constraints.
The performance of the ACHCI algorithm is dependent on the spatial characteristics of the precedence constraints, and it performs quite well when the child nodes are centrally located, outperforming the NN heuristic in 97\% of the studied instances.
Thus, this algorithm is highly applicable for certain configurations of pick up and delivery activities such as within a flexible manufacturing system that has inventory shelves at the perimeters of the facility.
Due to the constraints added to the TSP formulation, the ACHCI tour cost does not perform as well in cases where the precedence constraints have a random spatial characteristic or when the parent nodes are centrally located.

Future work will focus on extending the CHCI algorithm to consider the multi-commodity one-to-one pickup-and-delivery traveling salesperson problem and the dial-a-ride problem.

\section*{Acknowledgements}

This work was supported by Ford Motor Company through the Ford-OSU Alliance Program. Declarations of interest: none

\bibliography{cHull}

\end{document}